\documentclass[journal]{IEEEtran}

\ifCLASSINFOpdf
\else
   \usepackage[dvips]{graphicx}
\fi
\usepackage{url}

\usepackage{graphicx}
\usepackage{multirow}
\usepackage{colortbl}
\usepackage{xcolor}
\usepackage{bm}
\usepackage[pagebackref,breaklinks,colorlinks]{hyperref}
\usepackage{cite}
\usepackage{booktabs} 
\usepackage{adjustbox}
\RequirePackage{amsmath}
\RequirePackage{amssymb}

\begin{document}

\title{Dataset Distillation for Super-Resolution without Class Labels and Pre-trained Models}

\author{Sunwoo Cho, Yejin Jung, Nam Ik Cho, \IEEEmembership{Senior Member, IEEE}, and Jae Woong Soh, \IEEEmembership{Member, IEEE}
\thanks{Sunwoo Cho and Nam Ik Cho are with the Department of Electrical and Computer Engineering, INMC, Seoul National University, Seoul 08826, Republic of Korea (e-mail: etoo33@snu.ac.kr; nicho@snu.ac.kr).}
\thanks{Yejin Jung is with the Graduate School of Engineering Practice, INMC, Seoul National University, Seoul 08826, Republic of Korea (e-mail: yejin93.jung@snu.ac.kr).}
\thanks{Jae Woong Soh is with the Department of Electrical Engineering and Computer Science, Gwangju Institute of Science and Technology, Gwangju 61005, Republic of Korea (e-mail: jaewoongsoh@gist.ac.kr).}}

\markboth{Journal of \LaTeX\ Class Files, Vol. 14, No. 8, August 2015}
{Shell \MakeLowercase{\textit{et al.}}: Bare Demo of IEEEtran.cls for IEEE Journals}
\maketitle

\begin{abstract}

Training deep neural networks has become increasingly demanding, requiring large datasets and significant computational resources, especially as model complexity advances. Data distillation methods, which aim to improve data efficiency, have emerged as promising solutions to this challenge. In the field of single image super-resolution (SISR), the reliance on large training datasets highlights the importance of these techniques. Recently, a generative adversarial network (GAN) inversion-based data distillation framework for SR was proposed, showing potential for better data utilization. However, the current method depends heavily on pre-trained SR networks and class-specific information, limiting its generalizability and applicability. To address these issues, we introduce a new data distillation approach for image SR that does not need class labels or pre-trained SR models.  
In particular, we first extract high-gradient patches and categorize images based on CLIP features, then fine-tune a diffusion model on the selected patches to learn their distribution and synthesize distilled training images.
Experimental results show that our method achieves state-of-the-art performance while using significantly less training data and requiring less computational time. Specifically, when we train a baseline Transformer model for SR with only 0.68\% of the original dataset, the performance drop is just 0.3 dB. In this case, diffusion model fine-tuning takes 4 hours, and SR model training completes within 1 hour, much shorter than the 11-hour training time with the full dataset.

\end{abstract}


\IEEEpeerreviewmaketitle

\section{Introduction}
\label{sec:intro}

\IEEEPARstart{S}{ingle} Image Super-Resolution (SISR) aims to reconstruct high-resolution images from their low-resolution counterparts. With the progress of deep learning technologies, SISR methods have advanced considerably, employing various model architectures and training strategies~\cite{lim2017enhanced, wang2021real, zhou2023srformer, hossain2024lightweight, huang2024multi, wu2025adaptive, xiao2022heterogeneous}. As these models have grown more complex, their dependence on large-scale training datasets has also increased. This has led to the expansion of datasets, starting from 800 2K images in DIV2K~\cite{agustsson2017ntire}, to 2,650 images in Flickr2K~\cite{timofte2017ntire}, and further increasing to 10,324 images in OST~\cite{wang2018recovering}.

However, it is believed that not all images in large datasets contribute equally to model learning~\cite{ding2023not}. Some data points may offer little value while consuming significant computational resources. As datasets grow, demands on memory and processing power increase, making it difficult to train SR models efficiently. To address these issues, it is essential to develop effective dataset curation strategies that enhance model training while saving computational resources.


\begin{figure}[t]
\begin{center}
\includegraphics[width=0.9\linewidth ]{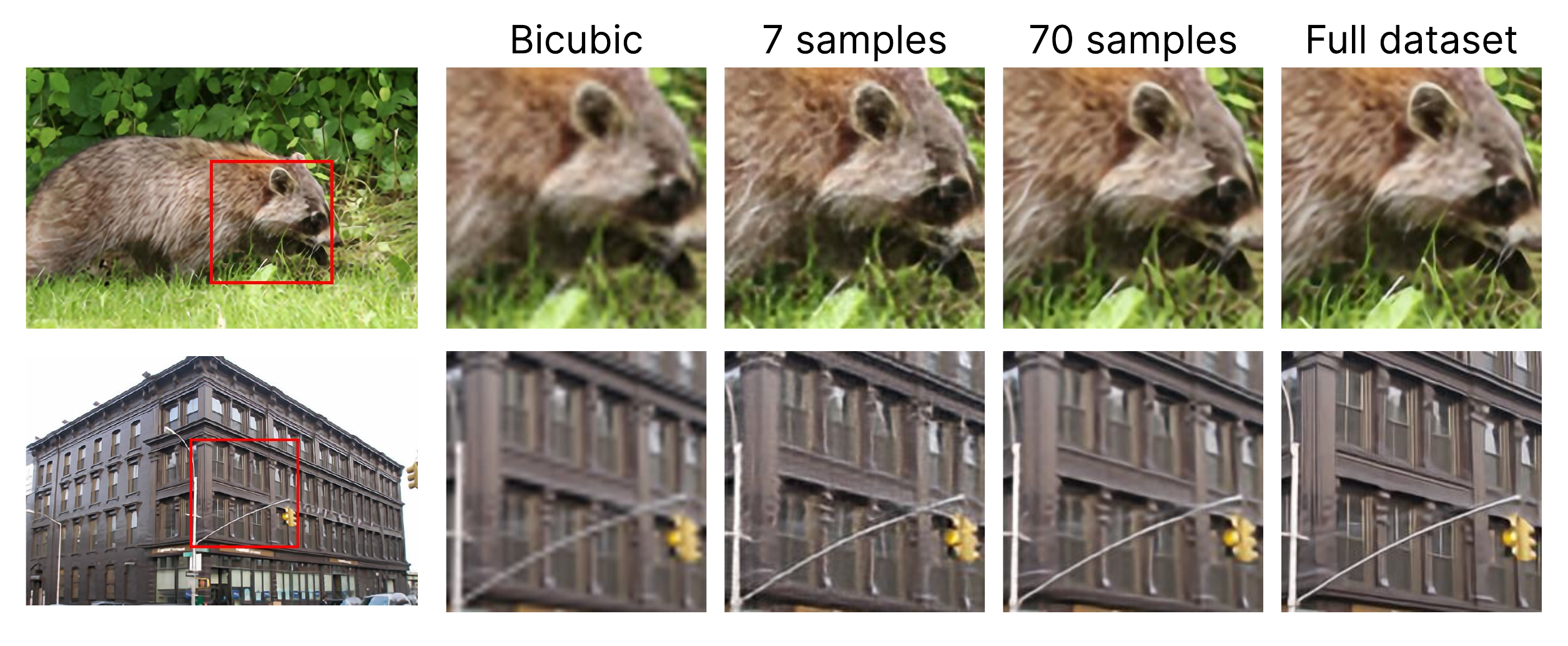}
\vspace{-10pt}
\caption{Qualitative SR comparison. A model trained on 70 distilled images matches the full-set baseline.}
\label{fig:SR_result}
\end{center}
\vspace{-25pt}
\end{figure}

In this context, coreset selection methods~\cite{katharopoulos2018not, phillips2017coresets, welling2009herding} aim to identify effective training subsets but are limited by dependence on the original dataset. To overcome this, Wang et al.~\cite{wang2018dataset} introduced dataset distillation, inspired by knowledge distillation~\cite{hinton2015distilling}, which compresses large datasets into a few synthetic samples optimized for downstream tasks. Subsequent studies refined objectives and architectures, improving generalization and stability \cite{lei2023comprehensive, yu2023dataset, zhao2020dataset, cazenavette2022dataset}. Recent work distills directly in latent space using GAN and diffusion priors \cite{goodfellow2020generative, ho2020denoising, su2025generative, cazenavette2023generalizing, su2024d}. Gu et al.~\cite{gu2024efficient} boost efficiency by fine-tuning a pretrained diffusion model with a minimax loss, producing distilled sets that train diverse networks faster without loss of accuracy.

In the field of SISR, Zhang et al.~\cite{zhang2024gsdd} introduced data distillation with their Generative Super-Resolution Dataset Distillation (GSDD) method.
This approach employs GAN inversion~\cite{xia2022gan} guided by a pre-trained SR model and optimized through feature matching to capture the original dataset's distribution in the latent space. This leads to significant efficiency gains, achieving a 93.2\% decrease in storage requirements.
Despite its strong results, GSDD has two main limitations. First, it depends on a pre-trained SR model to generate the distilled dataset, which causes inconsistent performance across different SR model architectures. Second, it requires class labels, which are often unavailable in SR datasets, thereby limiting its practical application.

\begin{figure*}[t]
\vspace{-10pt}
\begin{center}
\includegraphics[width=0.88\linewidth]{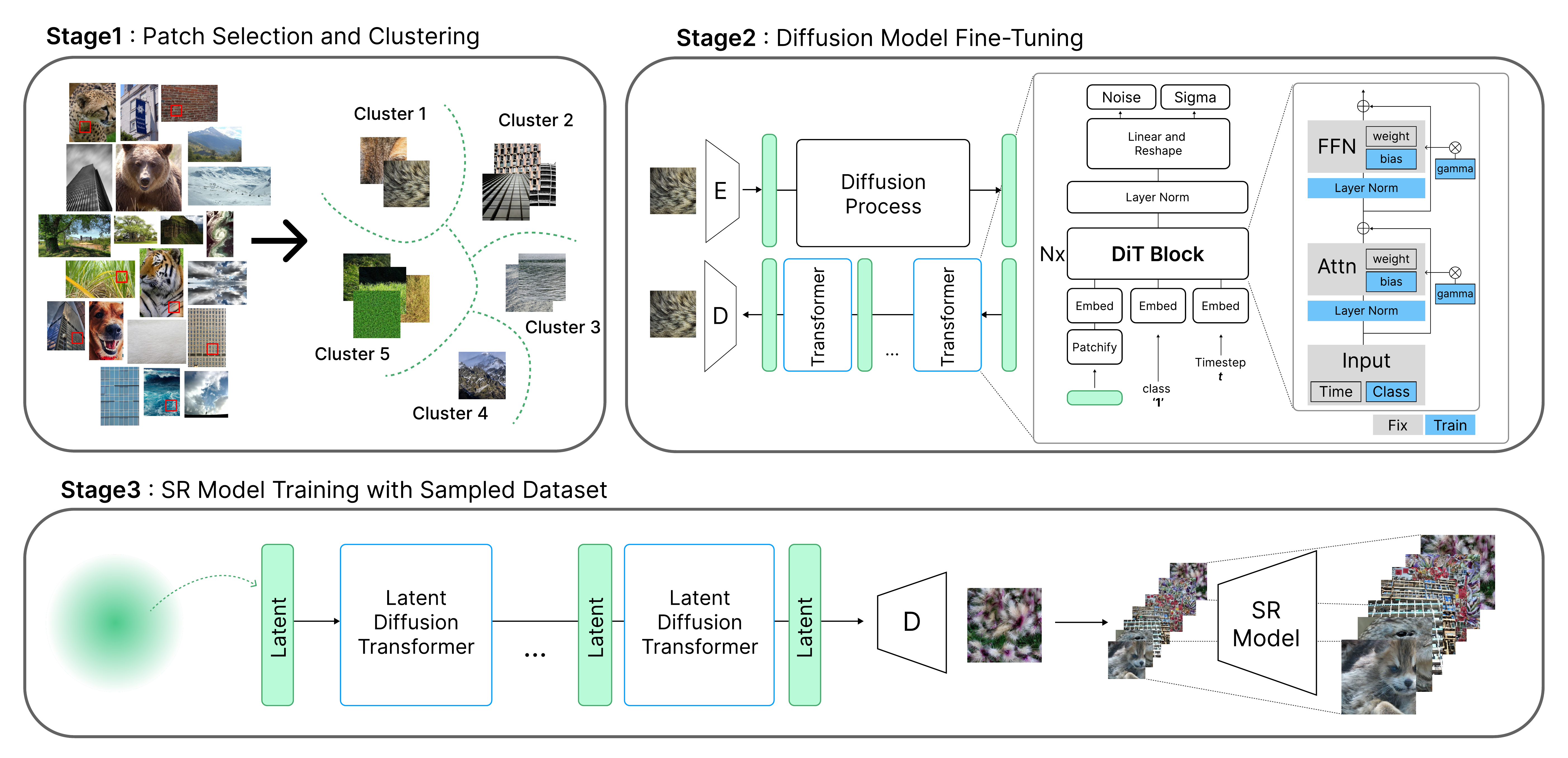}
\vspace{-15 pt}
\caption{Overview of our dataset-distillation process. }
\label{fig:Overview}
\end{center}
\vspace{-20pt}
\end{figure*}


To address the issues above, we introduce a novel data distillation framework for SISR that does not require pre-trained SR models or class labels. Our main idea is to replace explicit class labels with a semantic feature space and use a Latent Diffusion Model (LDM)~\cite{rombach2022high} to select informative training samples. Our method includes three main stages.
First, we implement an informative patch selection strategy combined with clustering to generate pseudo-labels. Second, we fine-tune the diffusion model using a composite loss that combines the Minimax loss function~\cite{gu2024efficient} with a loss specifically designed for SR. Finally, we generate synthetic training data from the fine-tuned diffusion model to train the SR network.

Extensive experiments demonstrate that our method delivers performance on par with existing approaches while eliminating the dependence on pre-trained SR networks and class-specific annotations. Additionally, it significantly shortens overall training time and requires much less storage compared to using the full dataset. We have conducted thorough experimental validation across various SR architectures to confirm the effectiveness and generalizability of our approach. Notably, when applied to SRFormer~\cite{zhou2023srformer}, our method achieves similar results with only a 0.3 dB decrease in PSNR, while using just 0.68\% of the original dataset. As shown in Fig.~\ref{fig:SR_result}, an SR model trained on only 70 distilled images produces results comparable to those trained on the full dataset.

\section{Method}
\label{sec:Method}

We propose a three-stage framework for efficient data distillation: (1) {\bf Patch Selection and Clustering}; (2) {\bf Diffusion Model Fine-Tuning}; and (3) {\bf SR Model Training with Sampled Dataset}. An overview is shown in Fig.~\ref{fig:Overview}.

\subsection{Patch Selection and Clustering}
To effectively distill SR datasets without using explicit class labels, we propose a systematic preprocessing pipeline for estimating pseudo-labels.
Our approach begins with a strategic patch selection to identify the most informative regions within the dataset by removing low-texture patches, which provide little information for SR.

In particular, we quantitatively evaluate the informativeness of each patch by calculating the PSNR between each high-resolution patch $X$ and its reconstructed version obtained through sequential downsampling and upsampling operations following \cite{wang2021samplingaug}, as $\mathcal{PSNR}_{bic}$ provides a simple and widely used measure of high-frequency information.

\begin{equation}\label{eq:simple}
    \mathcal{PSNR}_{bic}=PSNR(X,(X\downarrow_{bic4})\uparrow_{bic4}).
\end{equation}
where $\downarrow_{bic4}$ and $\uparrow_{bic4}$ denote bicubic downsampling and upsampling operations with a scale factor of 4, respectively. Patches with PSNR values below a predetermined threshold are selected as candidate patches, since lower PSNR indicates greater information loss through downsampling.

Next, to enhance the efficiency of fine-tuning the diffusion model, we propose a clustering-based pseudo-labeling strategy for the selected patches. Specifically, we extract distinct features using the CLIP model \cite{radford2021learning}, and apply $k$-means clustering to generate pseudo labels. This approach allows more structured and effective training without the need for explicit class annotations. The visualizations of the clustering results are shown in Fig.~\ref{fig:visualization}.


\begin{table*}[t]
    \centering
    \caption{Performance comparison with the state-of-the-art dataset distillation method for super-resolution. The best performance within the same model architecture is highlighted in \textbf{bold}.}
    \label{tab:sota_comparison}
    \vspace{-5pt}
      \scriptsize
  \setlength{\tabcolsep}{2pt}
  \renewcommand{\arraystretch}{0.9}
    \begin{adjustbox}{width=\textwidth}
    
    \begin{tabular}{@{}l|cccccc|c@{}}
        \toprule
        \multicolumn{1}{c|}{\textbf{Number of Images}}   & 
         \textbf{7} & \textbf{35} & \textbf{70} & \textbf{140} & \textbf{280} & \textbf{420} &  \textbf{FULL} \\ 
         \multicolumn{1}{c|}{\textbf{Ratio}} &  0.07\% & 0.34\% & 0.68\% & 1.36\% & 2.71\% & 4.07\% &  100\% \\ 
       \midrule 
        \textbf{Model Architecture }
       & SSIM$\uparrow$ / LPIPS$\downarrow$ & SSIM$\uparrow$ / LPIPS$\downarrow$ & SSIM$\uparrow$ / LPIPS$\downarrow$ & SSIM$\uparrow$ / LPIPS$\downarrow$ & SSIM$\uparrow$ / LPIPS$\downarrow$ & SSIM$\uparrow$ / LPIPS$\downarrow$ & SSIM$\uparrow$ / LPIPS$\downarrow$ \\

       \specialrule{1pt}{1.5pt}{1pt}
       \specialrule{1pt}{1.5pt}{1pt}

        \textbf{RealSR-GSDD}
         & 0.4173 / 0.5874 & 0.4323 / 0.5667 & 0.4288 / 0.5595 & 0.4042 / 0.4723 & 0.5142 / 0.4089 & 0.6205 / 0.3805 & \multirow{2}{*}{ 0.8254 / 0.1123 } \\ \cmidrule{1-7}

        \textbf{RealSR-Ours}
          
        & \textbf{0.5828} / \textbf{0.4164}  & \textbf{0.6095} / \textbf{0.3742} & \textbf{0.6289} / \textbf{0.3543} & \textbf{0.6335} / \textbf{0.3476} & \textbf{0.6302} / \textbf{0.3413} & \textbf{0.6389} / \textbf{0.3229}  \\ 

        \specialrule{1pt}{1.5pt}{1pt}
        \specialrule{1pt}{1.5pt}{1pt}
        

        \textbf{RealESRGAN-GSDD}
         & 0.4805 / 0.5102 & 0.4201 / 0.5029 & 0.4992 / 0.4807 & 0.4964 / \textbf{0.4162} & 0.5140 / \textbf{0.3841} & 0.6273 / \textbf{0.3512} & \multirow{2}{*}{0.8367 /  0.0903}   \\ 
        \cmidrule{1-7}


        \textbf{RealESRGAN-Ours}
        & \textbf{0.6269} / \textbf{0.4609} & \textbf{0.6323} / \textbf{0.4410} & \textbf{0.6299} / \textbf{0.4501} & \textbf{0.6292} / 0.4459 & \textbf{0.6294} / 0.4495 & \textbf{0.6451} / 0.4482 \\ 

        \bottomrule
    \end{tabular}
    \end{adjustbox}
    \vspace{-10pt}
\end{table*}

\begin{table*}[t]
    \centering
    \caption{Performance comparison in diverse super-resolution model architectures.}
    \label{tab:SRmodel_comparison}
    \vspace{-5pt}
          \scriptsize
  \setlength{\tabcolsep}{2pt}
  \renewcommand{\arraystretch}{0.9}
    \begin{adjustbox}{width=\textwidth}
    
    \begin{tabular}{@{}l|cccccc|c@{}}
        \toprule
        \multicolumn{1}{c|}{\textbf{Number of Images}}   & 
         \textbf{7} & \textbf{35} & \textbf{70} & \textbf{140} & \textbf{280} & \textbf{420} &  \textbf{FULL} \\ 
         \multicolumn{1}{c|}{\textbf{Ratio}} &  0.07\% & 0.34\% & 0.68\% & 1.36\% & 2.71\% & 4.07\% &  100\% \\ 
         
       \midrule 
        \textbf{Model Architecture }
       & PSNR$\uparrow$ / SSIM$\uparrow$ & PSNR$\uparrow$ / SSIM$\uparrow$ & PSNR$\uparrow$ / SSIM$\uparrow$& PSNR$\uparrow$ / SSIM$\uparrow$& PSNR$\uparrow$ / SSIM$\uparrow$ & PSNR$\uparrow$ / SSIM$\uparrow$ & PSNR$\uparrow$ / SSIM$\uparrow$ \\

       \midrule 

        \textbf{EDSR}
        & 25.03 / 0.7021 & 25.14 / 0.7089 & 25.21 / 0.7109 & 25.32 / 0.7177 & 25.38 / 0.7207 & 25.41 / 0.7219  & 25.72 / 0.7348  \\  \midrule

        \textbf{RCAN}
          
        & 24.89 / 0.6935 & 25.10 / 0.7085 & 25.17 / 0.7157 & 25.29 / 0.7161 & 25.35 / 0.7185 & 25.39 / 0.7212 & 25.73 / 0.7355   \\ 

        \midrule

        \textbf{SRFormer}
        & 26.15 / 0.6884 & 26.56 / 0.7109 &26.73 / 0.7157 & 26.77 / 0.7188 & 26.82 / 0.7219 & 26.85 / 0.7227 & 27.05 / 0.7297   \\

        \midrule

        \textbf{HiTSR}
& 26.17 / 0.6895 &26.48 / 0.7081& 26.68 / 0.7134 &	26.75 / 0.7167 &	 26.82 / 0.7205 & 26.82 / 0.7212 & 27.71 / 0.7548 \\ 

        \bottomrule
    \end{tabular}
    \end{adjustbox}
    \vspace{-10pt}
\end{table*}


\subsection{Diffusion Model Fine-Tuning }

In this step, we use pre-trained diffusion models to create distilled datasets for SR. By fine-tuning these models on original SR datasets, we generate compact yet representative SR datasets that maintain the original data distribution while improving training efficiency.

Diffusion models learn the data distribution by progressively corrupting images with Gaussian noise and then learning to reverse this process. In the Latent Diffusion Model framework~
\cite{rombach2022high}, the diffusion process is performed in a lower-dimensional latent space transformed by an encoder (E) and decoder (D). The diffusion learning objective is defined using a squared error loss function, mathematically expressed as: 
\begin{equation}\label{eq:simple}
    \mathcal{L}_{simple}=||\epsilon_\theta(\mathbf{z}_t,\mathbf{c})-\epsilon||^2_2,
\end{equation}
where $\epsilon$ denotes the true noise, and $\epsilon_{\theta}(\mathbf{z}_t,\mathbf{c})$ is the noise predicted from the latent representations $\mathbf{z}_t$ and optional class information $\mathbf{c}$ at timestep $t$.

For fine-tuning the diffusion model, we propose a novel loss function based on the Minimax loss introduced in \cite{gu2024efficient}. The Minimax loss enhances both the representativeness and diversity of generated samples by combining two complementary loss terms: $\mathcal{L}_{r}$ and $\mathcal{L}_{d}$.

First term: the representativeness loss,
\begin{equation}
\label{eq:lossrepresentative}
    \mathcal{L}_{r}=\arg\max_{\theta}\min_{m\in [M]}\sigma\left(\hat{\mathbf{z}}_\theta(\mathbf{z}_t,\mathbf{c}),\mathbf{z}_m\right),
\end{equation}
where $\hat{\mathbf{z}}_\theta(\mathbf{z}_t,\mathbf{c}) = \mathbf{z}_{t} - \epsilon_{\theta}(\mathbf{z}_{t},\mathbf{c})$ and $\sigma$ denotes the cosine similarity. 
We store a small memory $[M]=\{\,z_{m}\,\}_{m=1}^{N_{M}}$ of real latent features drawn in recent fine-tuning steps.
By maximizing the minimum cosine similarity between the predicted sample and the stored real samples, model pulls predicted samples toward the ``center'' that best covers all recently seen real features, ensuring global representatives.

Second term: the diversity loss,
\begin{equation}\label{eq:lossdiversity}
    \mathcal{L}_{d}=\arg\min_\theta\max_{d\in [D]}\sigma\left(\hat{\mathbf{z}}_\theta(\mathbf{z}_t,\mathbf{c}),\mathbf{z}_d\right),
\end{equation}
where we keep a sliding memory of recent synthetic latents $[D]=\{\,z_{d}\,\}_{d=1}^{N_{d}}$. We minimize the maximum cosine similarity between the current predicted latent to any previously generated latent samples. This loss helps to generate each new sample away from its closest peer, encouraging the distilled set to cover the latent space more uniformly and yielding more diverse samples.

Additionally, we propose adding a high-frequency-aware loss term to the Minimax loss, specifically designed for super-resolution applications. This loss is formulated as:

\begin{equation}\label{eq:lossdiversity}
    \mathcal{L}_{SR}=\arg\max_\theta||{\mathbf{\hat{x}_{bic}}}-\mathbf{\hat{x}}||^2_2,
\end{equation}
where $\mathbf{\hat{x}}$ represents the predicted high-resolution image, and $\mathbf{\hat{x}_{bic}}$ denotes the bicubic up-sampled low-resolution image. This term explicitly encourages the model to preserve and enhance high-frequency components in the generated images.

The overall training loss is expressed as:
\begin{equation}
\label{eq:total}
    \mathcal{L}=\mathcal{L}_{simple}+\lambda_r\mathcal{L}_r+\lambda_d\mathcal{L}_d+\lambda_{sr}\mathcal{L}_{SR},
\end{equation}
where $\lambda_r$, $\lambda_d$, and $\lambda_{sr}$ are hyperparameters controlling the weights of each loss term. During training, we employ pseudo-class labels derived from the initial clustering stage. This integrated objective function optimizes both the diffusion model's ability to learn the data distribution and its capacity to generate high-frequency details. As a result, it produces synthetic images that are especially effective for training SR models.

\subsection{SR Model Training with Sampled Dataset}

Finally, we sample images directly from a fine-tuned diffusion model in latent space. In contrast to GSDD~\cite{zhang2024gsdd}, which performs sample-wise iterative optimization, our approach directly samples datasets from the fine-tuned diffusion model. The sampled images serve as HR images and their bicubic  downsampled versions form LR counterparts for SR training. Since our framework does not depend on pre-trained SR models, the distilled data can be applied across various SR architectures. Additionally, our method significantly reduces overall training time and requires much less storage compared to using the full dataset.

\section{Experiments}
\label{sec:expeeeriments}

\subsection{Implementation Details}
\label{sec:implementation  details}
We conduct experiments using the Outdoor Scene Test (OST) dataset~\cite{wang2018recovering}, which contains 10,324 images across seven semantic categories: animals, plants, grass, buildings, sky, water, and mountains. We note that, unlike GSDD~\cite{zhang2024gsdd}, our approach uses only raw image data without category annotations. For evaluation, we use the OutdoorSceneTest300 benchmark (300 test images) with SSIM ~\cite{wang2004image} and LPIPS ~\cite{zhang2018unreasonable} as primary metrics, enabling direct comparison with GSDD. Additionally, we use PSNR and SSIM for cross-architecture validation of non-GAN-based models.

For patch selection, we set the $\mathcal{PSNR}_{bic}$ threshold at 23 dB, the median of the patch distribution, to balance the trade-off between the quantity and quality of patches. The method is stable across thresholds. For clustering, we adopt the CLIP~\cite{radford2021learning} features, specifically the ViT `Base' model with a patch size of 32~\cite{alexey2020image}, and set $k=7$, corresponding to the seven class labels in the OST dataset for fair comparison. Experiments indicate robustness to cluster number.

For our diffusion model architecture, we employ a Diffusion-based Image Transformer (DiT)~\cite{peebles2023scalable}, enhanced with DiffFit~\cite{xie2023difffit}, a parameter-efficient fine-tuning method. During both fine-tuning and inference, images are processed at a resolution of $512 \times 512$ pixels. The model is fine-tuned for 70 epochs with a batch size of 8, using the Adam optimizer~\cite{kingma2014adam} ($\beta_1 = 0.9, \beta_2 = 0.999$, with a learning rate of $1 \times 10^{-3}$). 

We set the joint loss weights to $\lambda_{r}= 0.002$, $\lambda_{d} = 0.008$, and $\lambda_{sr} = 1$, following Minimax Diffusion~\cite{gu2024efficient}. We found that performance is robust to these values. For the SR model training, we use publicly available implementations and retrain them on our distilled dataset, using bicubic downsampling to generate paired samples at a scale of $4\times$, consistent with GSDD~\cite{zhang2024gsdd}. Using a single NVIDIA RTX 3090 GPU, fine-tuning the diffusion model takes approximately 4 hours, and SR model training requires 1 hour, resulting in a total of 5 hours compared to 11 hours when training directly on the full dataset.

\subsection{Comparison with State-of-the-art Methods}

We conduct comprehensive evaluations of our distillation method using two state-of-the-art SR architectures: Real-ESRGAN~\cite{wang2021real} and RealSR~\cite{ji2020real}.
Table~\ref{tab:sota_comparison} shows performance metrics for different sizes of distilled training sets. The results demonstrate a positive relationship between the size of the distilled dataset and SR accuracy, attributable to the increased diversity of informative features in larger sample sets. Notably, our method achieves superior performance without depending on pre-trained SR models or class labels. The approach performs well, especially in extreme condensation scenarios, demonstrating the effectiveness of our distillation strategy.

\subsection{Cross-architecture Validation}
\label{sec:cross_arch}
We perform comprehensive evaluations across various SR architectures, including the residual-based model EDSR~\cite{lim2017enhanced}, attention-driven model RCAN~\cite{zhang2018image}, Transformer-based models SRformer-light~\cite{zhou2023srformer} and HiT-SR~\cite{zhang2024hierarchical}. The experimental results shown in Table~\ref{tab:SRmodel_comparison} demonstrate that our method maintains strong performance across these architectures. Additionally, as shown in Table~\ref{tab:sota_comparison}, the SR quality positively correlates with the number of distilled samples in the training set.

While existing methods like GSDD are limited to GAN-based architectures and heavily rely on the trained SR model, our approach shows exceptional versatility. It easily works with both residual networks and transformer architectures, regardless of model capacity. Notably, when applied to SRformer~\cite{zhou2023srformer}, our method achieves only a 0.3 dB reduction in PSNR, while utilizing just 0.68\% of the original training data. These results confirm the efficiency and effectiveness of our proposed approach.

\subsection{Ablation Studies}
\label{sec:ablation}
We perform ablation studies to assess the effectiveness of various dataset preparation strategies. Specifically, we compare (1) random crop, which uses uniform random sampling from the original dataset with fixed-size crops at arbitrary locations, (2) threshold crop selection, which applies PSNR-threshold-based filtering after random sampling and corresponds to stage 1 of our pipeline, and (3) our proposed distillation method. As shown in Table~\ref{tab:ablation}, the PSNR-threshold filtering approach consistently outperforms naive random selection across all dataset sizes. Additionally, our complete distillation framework achieves higher SSIM scores across all compression ratios, with especially notable improvements at higher compression rates. These findings confirm the effectiveness of our distillation strategy in maintaining image quality while greatly reducing dataset size.

\begin{table}[t]{
    \centering
    \caption{Quantitative comparison on Real-ESRGAN~\cite{ji2020real} trained on various different dataset condensation strategies.}
    \label{tab:ablation}
    \vspace{-5pt}
          \scriptsize
  \setlength{\tabcolsep}{2pt}
  \renewcommand{\arraystretch}{0.9}
    \centering
    \begin{adjustbox}{width=8.5cm}
    \centering
    \begin{tabular}{@{}lcccc@{}}
        \toprule
        \textbf{Number of Images}  & \textbf{140} & \textbf{280} & \textbf{420} \\ 
        \textbf{Ratio}  & 1.36\% & 2.71\% & 4.07\%  \\ 
        \textbf{Dataset type} & SSIM↑ / LPIPS↓ & SSIM↑ / LPIPS↓ & SSIM↑ / LPIPS↓ \\

       \midrule

        \multirow{1}{*}{\textbf{Random Crop}} 
        &  0.6068 / 0.4562 	
         & 0.6127 / 0.4449 & 0.6164 / 0.4445 \\ 

       \midrule
        \multirow{1}{*}{\textbf{Threshold Crop Select}}
         & 0.6251 / 0.4644 & 0.6269 / 0.4582 & 0.6139 / 0.4720 \\ 
         
       \midrule
        \multirow{1}{*}{\textbf{Our distilled dataset}} 
        & 0.6292 / 0.4459 &0.6294 / 0.4495 & 0.6451 / 0.4482 \\ 
        \bottomrule
    \end{tabular}
    \end{adjustbox}
   } 
   \vspace{-10pt}
\end{table}


\begin{figure}[t]
\begin{center}
\includegraphics[width=0.9\linewidth]{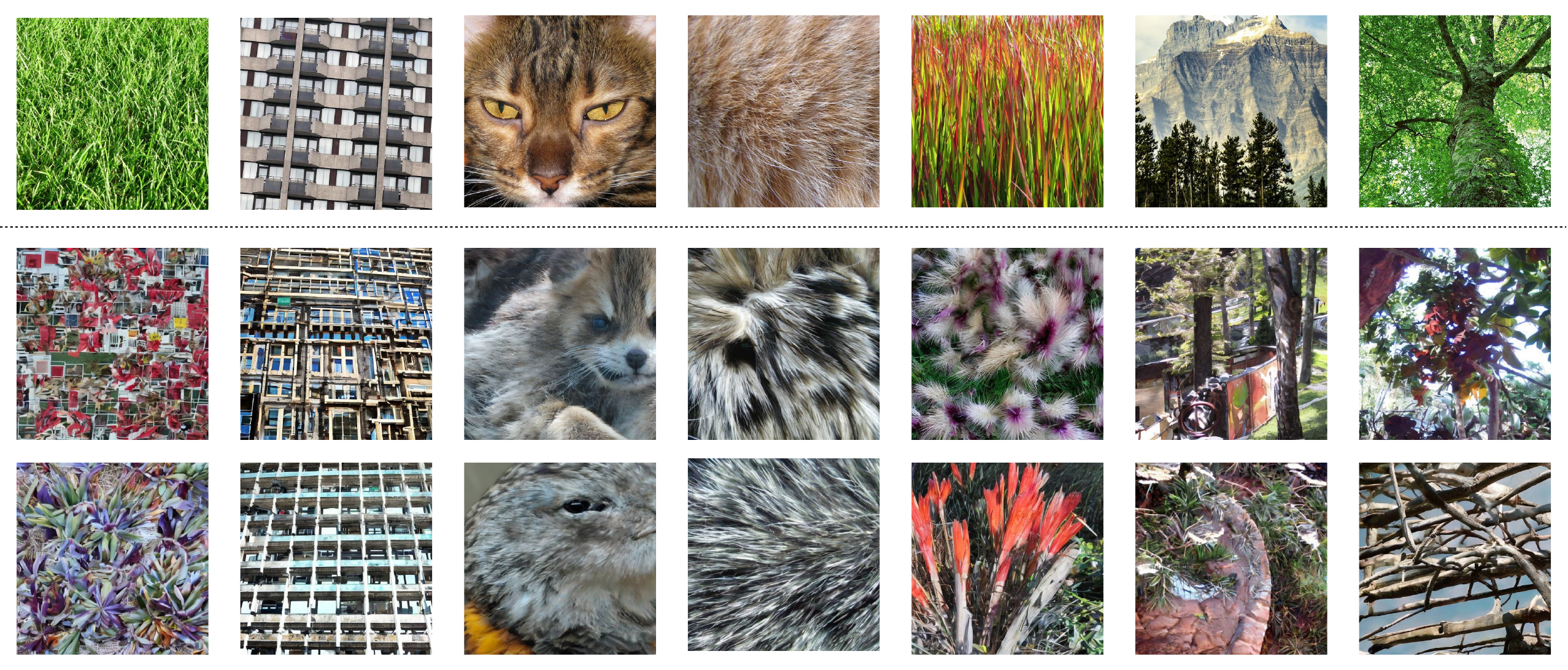}
\vspace{-10pt}
\caption{Visualization of our distilled datasets. First row presents real samples for each class, second and third row show the corresponding distilled images.}
\label{fig:visualization}
\end{center}
\vspace{-20pt}
\end{figure}

\subsection{Distilled Dataset Visualization}
\label{sec:ablation}
Fig. \ref{fig:visualization} illustrates the visualization of the distilled dataset. The top row presents representative samples from each initial cluster, illustrating distinct pseudo-label categories. The second and third rows display images synthesized by our distillation model, grouped by pseudo-class. These distilled images retain rich textures and align well with their respective classes, demonstrating that our method effectively captures and reproduces the original dataset’s complex textures while preserving semantic consistency.

\section{Conclusion}
\label{sec:majhead}

We presented a dataset distillation method for SR that requires neither class labels nor a pre-trained SR model. Our approach combines patch selection and clustering to form pseudo-class labels and applies a diffusion-based distillation strategy with an SR-specific loss. Once trained, the diffusion model can flexibly generate samples, and experiments on OST show superior and consistent performance across SR architectures.
However, performance improvements saturate beyond a certain number of distilled patches, and some hyperparameters may require dataset-specific tuning. Future work will focus on adaptive mechanisms to mitigate these limitations and further enhance scalability and robustness.

\vfill\pagebreak


\begingroup
\bibliographystyle{IEEEbib}
\bibliography{refs}
\endgroup

\end{document}